\documentclass[conference,10pt]{IEEEtran}
\IEEEoverridecommandlockouts
\usepackage{lineno}
\usepackage{cite}
\usepackage{amsmath,amssymb,amsfonts}
\usepackage{algorithmic}
\usepackage{graphicx}
\usepackage{textcomp}
\usepackage{xcolor}
\usepackage[numbers]{natbib}
\def\BibTeX{{\rm B\kern-.05em{\sc i\kern-.025em b}\kern-.08em
    T\kern-.1667em\lower.7ex\hbox{E}\kern-.125emX}}
\usepackage{microtype}

\usepackage{inconsolata}

\usepackage{booktabs} % for professional tables
\usepackage{tabularx}
\usepackage[pagebackref,breaklinks,colorlinks,citecolor=blue]{hyperref}

\usepackage{amsmath}
\usepackage{amssymb}
\usepackage{mathtools}
\usepackage{amsthm}
\usepackage{amsmath}

\usepackage{lipsum}  
\usepackage{listings}
\usepackage{caption}
\usepackage{subcaption}
\usepackage{CJKutf8}
\usepackage{xspace}
\usepackage{multirow}
\usepackage[framemethod=TikZ]{mdframed}
\usepackage{tabu}

\newcommand{\methodname}{$\text{CLAIR}_A$\xspace}

\renewcommand{\paragraph}[1]{\vspace{0.5em}\noindent\textbf{#1}$\:\:$}

\begin{document}

\title{\methodname: Leveraging Large Language Models to\\ Judge Audio Captions}

\author{\IEEEauthorblockN{Tsung-Han Wu, Joseph E. Gonzalez, Trevor Darrell, David M. Chan
\IEEEauthorblockA{\textit{Department of Electrical Engineering and Computer Science (EECS)} \\
\textit{University of California, Berkeley}\\
Berkeley, CA, USA \\
\texttt{\{tsunghan\_wu,jegonzal,trevordarrell,davidchan\}@berkeley.edu}}}\vspace{-1em}}

\maketitle

\begin{abstract}
Automated Audio Captioning (AAC) aims to generate natural language descriptions of audio. Evaluating these machine-generated captions is a complex task, demanding an understanding of audio-scenes, sound-object recognition, temporal coherence, and environmental context. While existing methods focus on a subset of such capabilities, they often fail to provide a comprehensive score aligning with human judgment. Here, we introduce \methodname, a simple and flexible approach that uses large language models (LLMs) in a zero-shot manner to produce a ``semantic distance'' score for captions. In our experiments, \methodname more closely matches human ratings than other metrics, outperforming the domain-specific FENSE metric by 5.8\% and surpassing the best general-purpose measure by up to 11\% on the Clotho-Eval dataset. Moreover, \methodname allows the LLM to explain its scoring, with these explanations rated up to 30\% better by human evaluators than those from baseline methods. The code for \methodname is made publicly available at \url{https://github.com/DavidMChan/clair-a}.

% The Automated Audio Captioning (AAC) task asks models to generate natural language descriptions of an audio input. Evaluating these machine-generated audio captions is a complex task that requires considering diverse factors, among them, auditory scene understanding, sound-object inference, temporal coherence, and the environmental context of the scene. While current methods focus on specific aspects, they often fail to provide an overall score that aligns well with human judgment. In this work, we propose \methodname, a simple and flexible method that leverages the zero-shot capabilities of large language models (LLMs) to evaluate candidate audio captions by directly asking LLMs for a semantic distance score. In our evaluations, \methodname better predicts human judgments of quality compared to traditional metrics, with a 5.8\% relative accuracy improvement compared to the domain-specific FENSE metric and up to 11\% over the best general-purpose measure on the Clotho-Eval dataset. Moreover, \methodname offers more transparency by allowing the language model to explain the reasoning behind its scores, with these explanations rated up to 30\% better by human evaluators than those provided by baseline methods. The code for \methodname is available at \url{https://anonymous.4open.science/r/clair-a-1666}.
\end{abstract}

\begin{IEEEkeywords}
Audio Captioning, Evaluation Metrics, Language Models, Auditory Scene Understanding
\end{IEEEkeywords}

\section{Introduction \& Background}

Audio captioning, generating a textual description for a sound, remains an ongoing and complex challenge in audio processing. Strong models designed for audio captioning must understand the sound and context wherein that sound occurs while expressing that information in natural language. A separate challenge, however, lies in evaluating the quality of these models. While the gold standard for evaluation is a human evaluation of caption quality \cite{drossos2017automated}, human evaluations are expensive and time-consuming. This expense indicates an imminent need to develop high-quality automated measures of caption quality that can be used to compare the semantic distance between human-written ground truth captions, and model-generated candidate captions. 

Often, approaches to audio captioning are evaluated with traditional natural language generation measures based on N-gram matching such as BLEU \cite{papineni2002bleu}, which counts the N-gram precision of the candidate sentence compared to a set of reference ground truths and ROUGE \cite{lin2004rouge}, which counts N-gram recall. A key issue with N-gram evaluation alone is that such measures cannot easily account for candidate sentences with identical semantic content to the references, but share few (if any) common N-grams. Some metrics were designed specifically to handle this issue: METEOR \cite{agarwal2008meteor} attempts to solve this problem with synonym-matching and stemming, and CIDEr \cite{vedantam2015cider} focused the n-gram matching on ``rare'' N-grams (using TF-IDF), as they are more likely to contain relevant semantic information.

\begin{figure}
    \centering
    \includegraphics[width=\linewidth]{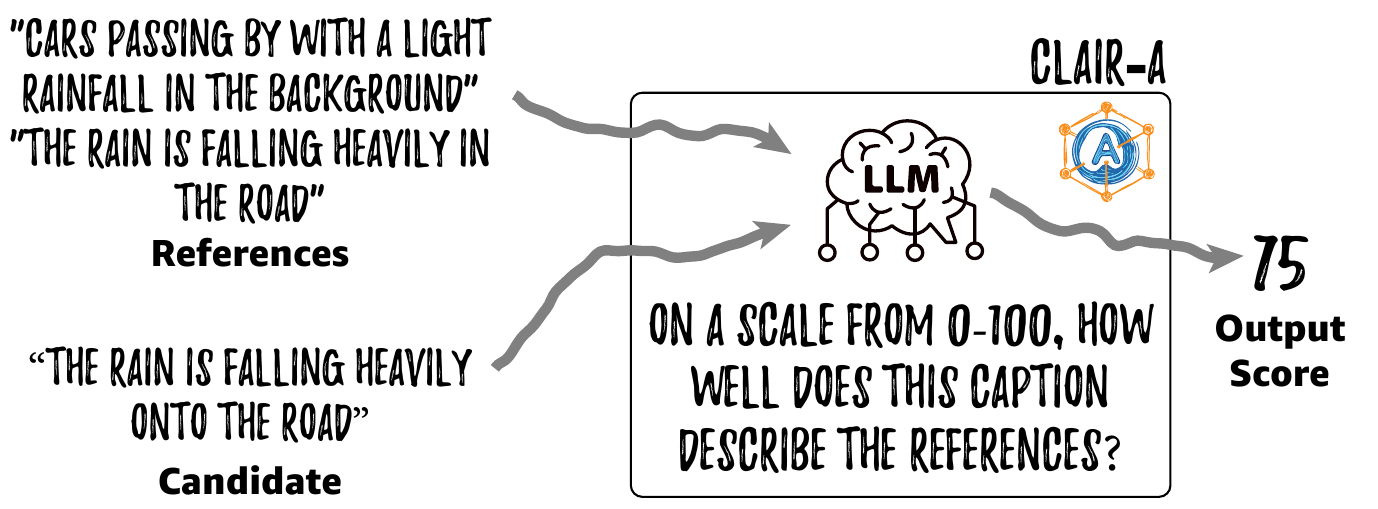}
    \vspace{0.25em}
    \caption{\methodname: a simple, domain-specific, measure for evaluating audio captions. By simply leveraging large language models, in-context learning, and guided generation techniques, \methodname both correlates strongly with human judgments of caption quality and produces both \textit{high-quality} and \textit{interpretable} scores according to human raters.}
    \label{fig:teaser}
\end{figure}

A key and prevailing idea among automated measures is that it is necessary to understand the ``relationships'' between objects in the scene (either objects in images or sound sources in audio captions). SPICE \cite{anderson2016spice} used the idea that image captions should parallel visual content by constructing ``object-graphs'' from parses of the captions, and comparing the ground truth object graphs with the candidate object graphs. SPIDEr \cite{liu2017improved}, a linear combination of SPICE and CIDEr, further aims to improve the improve the robustness of these measures. 

On the other hand, some measures have followed the thesis that such semantic similarity is inherent in the structure of language models. BERT-SCORE \cite{bert-score} and Sentence-BERT \cite{reimers-2019-sentence-bert} encode candidate and reference sentences as vectors using large language models, and compute distances between these vectors to produce a final semantic similarity. The most prevalent current audio captioning measure, FENSE \cite{zhou2022can}, extends this idea with an additional auxiliary score for local fluency detection to improve the robustness of the measure to non-fluent, but semantically similar generated captions. 

Some methods have aimed to combine the two approaches in a two-stage framework: SPICE+ \cite{gontier2023spiceplus} and ACES \cite{wijngaard2023aces} are both audio-captioning specific measures which first use a parser to extract either a parse graph (SPICE+) or explicit sound descriptors (ACES), and then use sentence-embedding methods to compare the resulting parses. With large language models (LLMs) such as GPT-4 \cite{openai2024gpt4o} showing promising results in the parsing space, the recently introduced X-ACE \cite{wang2024x} replaces many of the fixed components in SPICE with LLM-based parsers, and shows that the dynamic flexibility of LLMs can easily help extend some of the introduced rigidity in traditional domain-specific measures.

In this work, we go beyond such two-stage methods, and present \methodname, a novel, single-stage, approach that takes a highly simplified view of combining parsing and similarity. Inspired by recent work in image captioning \cite{chan2023clair}, and visual-question-answering \cite{bubeck2023sparks,dettmers2023qlora,vicuna2023}, instead of explicitly parsing the sentences, and then using semantic measures on the resulting parse, \methodname asks an LLM to score the semantic similarity between a candidate caption and reference set directly. By simply asking LLMs to produce a numeric score using in-context learning \cite{brown2020language}, \methodname aims to leverage already strong correlations with human judgment present in the base language models to solve semantic tasks without significant structural oversight. In addition to providing a score, we further ask the LLM to justify its answer in natural language. This justification is a unique benefit of \methodname, which allows the numeric score to be introspectable, leading to a measure that is directly human-interpretable. The code for \methodname is made publicly available at \url{https://github.com/DavidMChan/clair-a}.

Our key contributions are summarized as follows:
\begin{itemize}
    \item We introduce the \methodname measure, a simple and interpretable measure for audio captioning evaluation.
    \item We demonstrate that \methodname correlates better with human judgment than existing measures (both general and domain-specific), achieving up to 5.8\% relative accuracy improvement over the domain-specific FENSE metric and up to 11\% improvements over the best general-purpose measure on the Clotho-Eval dataset.
    \item We show that \methodname is interpretable in human judgment: humans rate the justifications generated by \methodname to be up to 30\% higher quality than na\"ive baselines.
\end{itemize}

\section{\methodname: LLMs as a Judge for Audio Captions}
\label{sec:method}
Given a candidate audio caption $c$, and a set of ground truth audio captions $G$, we would like to develop a score $S(c, G) \in [0, 1]$ which accurately predicts the semantic distance between $c$ and $G$. \methodname is inspired by CLAIR \cite{chan2023clair} (\underline{C}riterion using \underline{LA}nguage models for \underline{I}mage caption \underline{R}ating), and similarly leverages in-context-learning \cite{brown2020language} to convert audio caption evaluation to a text-completion task, which is solved using an off-the-shelf large language model (LLM), here, GPT-4o \cite{openai2024gpt4o}. The prompt, given in \autoref{fig:prompt}, encourages the large language model to produce a JSON output containing both (1) a numeric score between 1 and 100, and (2) a reason justifying that score, to provide interpretability. The numeric output of the LLM is used to generate the normalized LLM score:
\begin{equation}
    LLM(c, G) = \frac{\text{LLM result (0-100)}}{100}
\end{equation}

To ensure that the LLM produces a valid JSON output, we leverage efficient guided generation introduced in \cite{willard2023efficient}, which reformulates the text generation process of a standard LLM (which is usually done using temperature sampling from the likelihood distribution) by using a context-free grammar (CFG) to constrain the sampling process and ensure that sampled tokens conform to a valid JSON specification. A simple approach to this: checking each valid generated token for conformance to the CFG, and then re-sampling with that token masked if invalid, is prohibitively expensive because of LLMs' large vocabulary size and repeated evaluations of invalid tokens. To fix this, \cite{willard2023efficient} first construct a pushdown automaton parser for the grammar, and for every potential stack state of the parser, leverage pre-processing to pre-compute the valid next sampling tokens. These pre-computed token masks can then be efficiently queried (using a trie) at sampling time, with only one query needed per new token generated, guaranteeing that the next token that is generated by the LLM will be a valid continuation of the CFG. 

Unlike CLAIR, which uses re-sampling if the model generates errors, such an approach, which we implement using the Outlines library \cite{willard2023efficient}, guarantees valid parsing, and is significantly more efficient than CLAIR when handling invalid JSON generations. Another benefit over the re-sampling is that this allows \methodname to be fully deterministic (given a fixed LLM) when the sampling process is constrained by underlying CFG and is sampled with temperature zero, a key property for an automated measure.

\begin{figure}[t]
\centering
\noindent\begin{minipage}{\linewidth}
\mdfsetup{%
   % middlelinecolor=green,
   middlelinewidth=1pt,
   backgroundcolor=blue!3,
   innerleftmargin=0.5cm,
   innerrightmargin=0.5cm,
   roundcorner=15pt}
\begin{mdframed}
\vspace{0.2em}
\texttt{\small You are tasked with evaluating if a set of candidate captions accurately describes the same sound in a video clip as a reference set of captions. Start by assessing the accuracy and precision of how the audio characteristics are captured in the captions, scoring from 0 to 90 based on this aspect alone. After this initial assessment, you may add additional points (from 0 to 10) based on the quality of grammar and the detailed, reasonable descriptions present in the captions.}\vspace{0.5em}\\
\texttt{\small Candidate set:} \\
\textcolor{gray!70}{\texttt{\small \{candidate captions\}}} \\
\texttt{\small Reference set:} \\
\textcolor{gray!70}{\small \texttt{\{reference captions\}}} \vspace{0.5em}\\
\texttt{\small Combine these two aspects for a final evaluation score on a scale from 0 to 100, reflecting the likelihood that the candidate set is describing the same sound as the reference set. Format your response in JSON with a key "score", value between 0 and 100, and a key "reason" with a string value explaining your assessment.}
\end{mdframed}
\end{minipage}
\vspace{0.25em}
\caption{The prompt used for \methodname. Instead of asking for a single score, we find that a multi-tiered scoring system, which allocates points on a rubric, can mitigate ties and improve correlation on low-quality samples in the audio domain.}
\label{fig:prompt}
\vspace{-1em}
\end{figure}

Compared to recent measures such as X-ACE \cite{wang2024x}, SPICE+ \cite{gontier2023spiceplus} and ACES \cite{wijngaard2023aces}, which require a multi-step process that leverages LLMs or fixed parsers to transform captions into audio graphs which are then used for graph-matching (across sound events, sources, attributes, relationships, etc. either with LLMs or semantic vectors), \methodname is a simple, highly interpretable, zero-shot, approach which is easily transferable between languages (See \autoref{tab:multilingual}). 

While the LLM score alone can be powerful for distinguishing semantically varied captions (\autoref{tab:clotho}, \autoref{tab:audioset}), we found that in practice, many correct human captions are quite nuanced and similar, while many machine-generated audio captions are of poor quality, resulting in them receiving identical scores when assessed independently by the LLM. While this is not a problem for evaluating methods, it can be a problem when developing methods, as such tying scores cannot densely provide information to a researcher about which approaches are incremental improvements over others. To avoid ambiguities when the base LLM score is insufficient for distinguishing between competing candidates, we augment the base LLM score with an additional tie-breaking measure, yielding the final \methodname score:
\begin{equation}\label{eq:clair_a}
\text{CLAIR}_{A}(c, G) = LLM(c, G) + \epsilon \Gamma(c, G)
\end{equation}
where $\Gamma: (c, G) \to [0,1]$ is a normalized tie-breaking function and $\epsilon$ is a small constant (e.g., $\epsilon = 0.0001$). In \autoref{sec:results}, we consider several distinct choices for $\Gamma$, each introducing a different form of inductive bias or randomness: 
\begin{itemize}
    \item Random Tie-Breaking (\texorpdfstring{$\Gamma(c, G) \sim \text{Unif}(0,1)$}{Gamma ~ Unif(0,1)}): As a simple baseline, we set $\Gamma(c, G)$ to be a sample from the uniform distribution on $[0, 1]$. 
    \item Sentence-BERT Similarity: Alternatively, we use sentence-BERT \cite{reimers-2019-sentence-bert} to compute a semantic similarity score between the candidate $c$ and the reference set $G$. Specifically, $\Gamma(c, G)$ is set to the normalized cosine similarity between the sentence-BERT embeddings of $c$ and $G$. This leverages the representational power of pre-trained transformers to provide a more semantically informed tie-breaker.
    \item FENSE\cite{zhou2022can}: By setting $\Gamma(c, G)$ to the normalized FENSE score, we introduce a task-specific semantic tie-breaker designed explicitly for audio captioning.
\end{itemize}

We show in \autoref{sec:results} that incorporating these tie-breaking measures significantly improves performance, particularly for cases where multiple candidates receive similar LLM scores. Even with a very small $\epsilon$ ($= 0.0001$), the addition of $\Gamma$ helps distinguish between otherwise indistinguishable candidates, leading to more consistent and reliable evaluation outcomes.

% To avoid this, we augment the base LLM score with an additional tie-breaking measure to get the final \methodname score:
% \begin{equation}\label{eq:clair_a}
%     \text{CLAIR}_{A}(c, G) = LLM(c, G) + \epsilon \Gamma(c, G)
% \end{equation}
% where $\Gamma: (c, G) \to [0,1]$ is a normalized tie-breaking method. In \autoref{sec:results}, we discuss several choices for $\Gamma$ including $\Gamma(c, G) \sim \text{Unif}(0,1)$ (random), sentence-BERT and FENSE, and show that this significantly improves performance for samples that are either equally good or bad, even with $\epsilon = 0.0001$. 

Following experiments in \autoref{tab:tiebreaking}, we choose FENSE as a tie-breaking method with $\epsilon=0.25$ for the reference implementation.  Similar to \cite{chan2023clair}, we also consider a variant, $\text{CLAIR}_{AE}$, which averages across several LLMs to generate a mean LLM score, which is then summed with $\Gamma(c, G)$. This simple ensemble approach takes into account several LLM choices, which can often encode different aspects of human judgment.

\section{Results \& Discussion}
\label{sec:results}

\begin{figure*}
    \centering
    \includegraphics[width=0.95\linewidth]{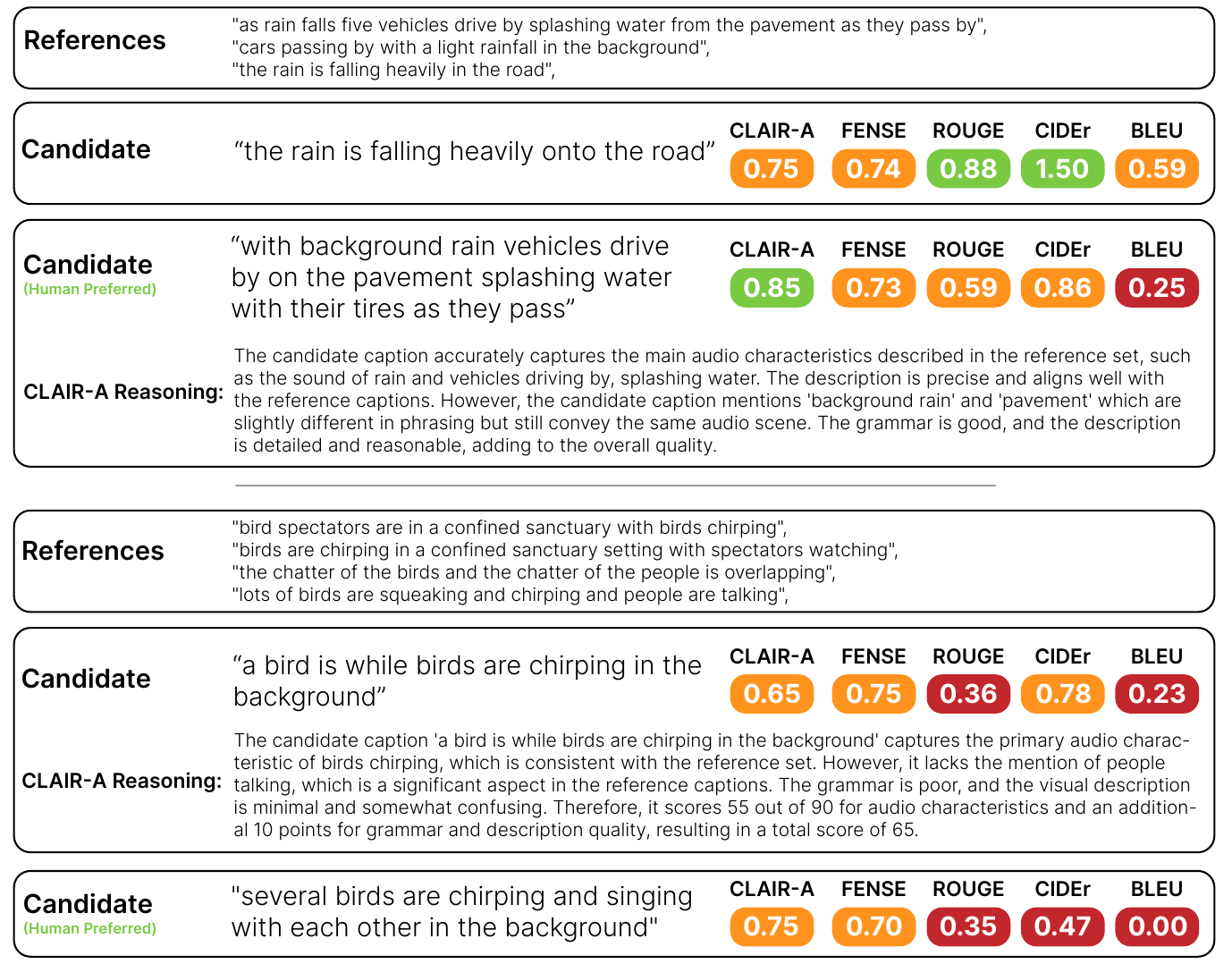}
    \caption{Qualitative examples of \methodname on the Clotho-Eval dataset. \methodname correlates strongly with human judgments while simultaneously giving detailed descriptions of \textit{why} such as judgment is reasonable. }
    \label{fig:qualitative}
\vspace{-1em}
\end{figure*}

To validate the performance of the \methodname measure, we perform several experiments targeting different aspects of the measure, including the correlation of the measure with human judgment, the performance on multilingual data, and the quality of the interpretable reasoning behind each of the caption scores. We benchmark against both standard measures of text similarity (BLEU \cite{papineni2002bleu}, METEOR \cite{agarwal2008meteor}, CIDEr \cite{vedantam2015cider}, SPICE \cite{anderson2016spice}, and CLAIR \cite{chan2023clair}) and specialized measures for audio captioning (SPIDEr \cite{liu2017improved}, Spice+ \cite{gontier2023spiceplus}, FENSE \cite{zhou2022can}, ACES \cite{wijngaard2023aces} and X-ACE \cite{wang2024x}).

\paragraph{Human Judgment:} Following \cite{zhou2022can}, we evaluate our measure on two datasets of pairwise human annotations for caption evaluation: the Clotho dataset \cite{drossos2020clotho} and the Audio-Caps dataset \cite{kim2019audiocaps}. These datasets, created by \cite{zhou2022can}, consist of 1,671 and 1,750 pairs of audio captions on Clotho and Audio-Caps respectively, with each pair of candidate captions annotated with ground truth reference captions, and human judgments of which caption better fits the ground truths. On this benchmark, the goal of a metric is to indicate reliably which caption is preferred by human raters, and we report the pair accuracy (a pair is ``correct'' if the preferred caption is assigned higher score).

Mirroring the design of \cite{vedantam2015cider}, tests are split into four categories: HC, which contains \textit{two correct human captions} describing the source audio, HI, which contains \textit{one correct, and one known incorrect human-generated caption} for the source audio, HM, which contains \textit{one correct human-generated caption, and one machine-generated caption} for the source audio, and MM which contains \textit{two machine-generated captions} for the source audio. Note in the HM and MM cases, it is not known if the machine-generated captions are correct or incorrect, rather, they were generated by a system to match the corresponding source audio. 

The accuracy of the metrics on each of the categories (HC, HI, HM, and MM), along with a total aggregate accuracy (mean micro-average), are shown for Clotho in \autoref{tab:clotho} and Audio-Caps in \autoref{tab:audioset}. We can see that \methodname outperforms other measures in all categories, with dramatic improvements in the HM and MM categories. It is worth noting that even though X-ACE leverages additional audio similarity in addition to the text content, \methodname still outperforms X-ACE overall, and significantly outperforms X-ACE without the cross-modal component. Since X-ACE only reports results on Clotho-Eval, we do not include its numbers for Audio-Caps. It is also clear that domain specialization for the measure is necessary. CLAIR alone, which is designed for image captioning, achieves only a 62.3\% total accuracy, demonstrating the necessity of per-domain specialization.

\begin{table}[t]
\small
\caption{Human preference match accuracy $(\uparrow)$ on the Clotho-Eval dataset. \methodname demonstrates significant improvement over both NLP and domain-specific measures. }
\label{tab:clotho}
\begin{center}
\begin{small}
\setlength{\tabcolsep}{4pt}
\begin{tabularx}{\linewidth}{Xccccc}
\toprule
Measure & HC & HI & HM & MM & All \\
\midrule
BLEU@1 \cite{papineni2002bleu}            & 51.0  & 90.6  & 65.5  & 50.3  & 59.0  \\
BLEU@4  \cite{papineni2002bleu}          & 52.9  & 88.9  & 65.1  & 53.2  & 60.5  \\
METEOR  \cite{agarwal2008meteor}         & 54.8  & 93.0  & 74.6  & 57.8  & 65.4  \\
ROUGEL \cite{lin2004rouge}          & 56.2  & 90.6  & 69.4  & 50.7  & 60.5  \\
CIDEr  \cite{vedantam2015cider}          & 51.4  & 91.8  & 70.3  & 56.0  & 63.2  \\
SPICE  \cite{anderson2016spice}          & 44.3  & 84.4  & 65.5  & 48.9  & 56.3  \\
BERTScore \cite{bert-score}        & 57.1  & 95.5  & 70.3  & 61.3  & 67.5  \\
% BLEURT           & 59.0  & 93.9  & 75.4  & 67.4  & 71.6  \\
Sentence-BERT \cite{reimers-2019-sentence-bert}    & 60.0  & 95.5  & 75.9  & 66.9  & 71.8  \\
CLAIR  \cite{chan2023clair}          & 42.9	& 95.9	& 72.8	& 54.8	& 62.3  \\
\midrule
SPICE+ \cite{gontier2023spiceplus} & 46.7 & 88.1 & 70.3 & 48.7 & 57.8 \\
ACES \cite{wijngaard2023aces}  & 56.7 & 95.5 & 82.8 & 69.9 & 74.0 \\
SPIDEr \cite{liu2017improved} & 53.3 & 93.4 & 70.3 & 57.0 & 64.2 \\
FENSE \cite{zhou2022can}   & 60.5  & 94.7  & 80.2  & 72.8  & 75.7  \\
\midrule
\methodname \\
$\quad$+ GPT-4o \cite{openai2024gpt4o} & \textbf{62.4} & \textbf{97.1} & \textbf{83.6} & \textbf{77.9} & \textbf{79.7} \\
$\quad$+ Gemini v1.5 (pro) \cite{gemini} & 59.0 & 95.9 & 83.2 & 75.1 & 77.4 \\
% $\quad$+ Llama 3.1 (8B) &    61.9 & 96.3 & 80.6 & 71.8 & 75.6 \\
$\quad$+ Phi Mini (3.5B) \cite{phi3} &  61.4 & 95.1 & 82.3 & 75.0 & 77.4 \\
CLAIR$_{AE}$ & 61.9	& \textbf{97.1} & 81.9 & 77.1 &	78.9 \\
\bottomrule
\end{tabularx}
\end{small}
\end{center}
\end{table}

\begin{table}[!ht]
\small
\caption{Human preference match accuracy $(\uparrow)$ on the AudioCaps-Eval dataset. \methodname can even outperform metrics augmented with cross-modal similarity such as X-ACE. }
\label{tab:audioset}
\begin{center}
\begin{small}
\setlength{\tabcolsep}{4pt}
\begin{tabularx}{\linewidth}{Xccccc}
\toprule
Measure & HC & HI & HM & MM & All \\
\midrule
BLEU@1 \cite{papineni2002bleu}          & 58.6  & 90.3  & 77.4  & 50.3  & 62.4  \\
BLEU@4 \cite{papineni2002bleu}          & 54.7  & 85.8  & 78.7  & 50.6  & 61.6  \\
METEOR \cite{agarwal2008meteor}          & 66.0  & 96.4  & 90.0  & 60.1  & 71.7  \\
ROUGEL \cite{lin2004rouge}          & 61.1  & 91.5  & 82.8  & 52.1  & 64.9  \\
CIDEr \cite{vedantam2015cider}            & 56.2  & 96.0  & 90.4  & 61.2  & 71.0  \\
SPICE \cite{anderson2016spice}            & 50.2  & 83.8  & 77.8  & 49.1  & 59.7  \\
BERTScore \cite{bert-score}       & 60.6  & 97.6  & 92.9  & 65.0  & 74.3  \\
% BLEURT           & \textbf{77.3}  & 93.9  & 88.7  & 72.4  & 79.3  \\
Sentence-BERT \cite{reimers-2019-sentence-bert}    & 64.0  & 99.2  & 92.5  & 73.6  & 79.6  \\
CLAIR \cite{chan2023clair}            & 44.8  & 99.2  & 90.0  & 56.4  & 67.4  \\
\midrule
SPICE+ \cite{gontier2023spiceplus} & 59.1 & 85.4 & 83.7 & 49.0 & 62.0 \\
ACES \cite{wijngaard2023aces} & 64.5 & 95.1 & 89.5 & 82.0 & 83.0 \\
SPIDEr \cite{liu2017improved} & 56.7 & 93.4 & 70.3 & 57.0 & 64.2 \\
FENSE \cite{zhou2022can}           & 64.5  & 98.4  & 91.6  & \textbf{84.6}  & 85.3  \\
X-ACE \cite{wang2024x}  & 69.7 & \textbf{99.6} & 93.7 & 76.8 & 81.8 \\
X-ACE w/o. CM \cite{wang2024x}  & 64.7 & 94.3 & 91.6 & 72.6 & 78.2 \\
\midrule
\methodname \\
$\quad$+ GPT-4o \cite{openai2024gpt4o} & 70.9	& 99.2	& 93.3 & \textbf{84.6}	& \textbf{86.6} \\
$\quad$+ Gemini v1.5 (pro) \cite{gemini} & 70.4 &	99.2 &	93.7 &	81.5 &	84.9 \\
% $\quad$+ Llama 3.1 (8B) &    66.0 & 98.4	& 92.9	& 68.9& 77.3 \\
$\quad$+ Phi Mini (3.5B) \cite{phi3} &  70.0 & 98.0 & \textbf{94.1} & 80.7 & 84.3 \\
CLAIR$_{AE}$ & \textbf{72.4} & \textbf{99.6} & 93.3 & 81.5 & 85.2 \\
\bottomrule
\end{tabularx}
\end{small}
\end{center}
% \vspace{-1em}
\end{table}

\paragraph{Multilingual Evaluation:} While most research in audio captioning is restricted to the English language, it is important to develop measures that transfer efficiently and effectively to multiple languages. To evaluate the performance of methods on multilingual data, we leveraged GPT-4o \cite{openai2024gpt4o} to translate the Clotho dataset to Chinese, and we retained the human annotations from the English language datasets. We then evaluate metrics zero-shot on the newly translated dataset and report their performance. Note that for \methodname, we explore two variants, a zero-shot variant where the prompt is un-translated (remains in English), and a language-aware variant, where the prompt is translated to the target language. We also leverage Sentence-BERT tiebreaking (as FENSE is incompatible with other languages). Our results are given in \autoref{tab:multilingual}, where we can see that \methodname translates flexibly to new languages with minimal or no adaptation and with minimal loss of accuracy, specifically for the HC cases. 

\begin{table}[t]
\caption{Human preference match accuracy $(\uparrow)$ on Clotho-Eval (Chinese). Multilingual BERTScore/Sentence-BERT/BLEU scores are used in this experiment.}
\label{tab:multilingual}
\begin{center}
\begin{small}
\setlength{\tabcolsep}{4pt}
\begin{tabularx}{\linewidth}{Xccccc}
\toprule
Measure & HC & HI & HM & MM & All \\
\midrule
BLEU@1 & 50.0 & 91.0 & 70.3 & 57.1 & 63.4 \\
BERTScore & 53.3 & 95.9 & 71.6 & 59.5 & 66.2 \\
Sentence-BERT & 56.2 & 93.9 & 78.9 & 66.6 & 71.3 \\
\midrule
\methodname  & 61.9 & \textbf{96.3} & 77.6 & 70.8 & 74.5 \\
\methodname (Language Aware) & \textbf{61.9} & 95.5 & \textbf{82.3} & \textbf{75.6} & \textbf{77.9} \\
\bottomrule
\end{tabularx}
\end{small}
\end{center}
\end{table}

\paragraph{Tie-Breaking:} One of the primary issues with the original CLAIR measure is the propensity of the method to generate ties when faced with equally good or bad data (which can be seen in the HC and MM column in \autoref{tab:audioset} and \autoref{tab:clotho}). This is a common problem for LLM-as-a-judge settings, where models often produce tie scores due to the discrete and coarse-grained scoring. Indeed, in these columns, the model generates a tying score of zero over 31\% of the time, leading to poor correlation. Thus, in \autoref{eq:clair_a}, we add an additional tie-breaking score to avoid inconclusive decisions. In \autoref{tab:tiebreaking} we demonstrate the performance of several tie-breaking methods. We can see that any tie-breaking method (including random) significantly improves the performance of the method, with ``intelligent'' tie-breaking methods leading to marginal improvements.

\begin{table}
\caption{Ablation of tie-breaking approaches on Clotho-Eval. \methodname (GPT-4o) used in all variants. The table shows human preference match accuracy $(\uparrow)$.}
\label{tab:tiebreaking}
\begin{center}
\begin{small}
\setlength{\tabcolsep}{4pt}
\begin{tabularx}{\linewidth}{Xccccc}
\toprule
Measure & HC & HI & HM & MM & All \\
\midrule
None & 42.4 &	96.3 &	75.9 &	64.7 &	68.3 \\
Random & 58.6 &	\textbf{97.1} &	82.3 &	74.7 &	77.6 \\
Sentence-BERT, $\epsilon=0.25$ & 61.4 &	\textbf{97.1} &	83.2 &	76.4 &	78.6 \\
FENSE, $\epsilon=0.0001$ & 61.9 &	\textbf{97.1} &	83.2 &	77.3 &	79.2 \\
FENSE, $\epsilon=0.25$ & \textbf{62.4} &	\textbf{97.1} &	\textbf{83.6} &	\textbf{77.9} &	\textbf{79.7} \\
\bottomrule
\end{tabularx}
\end{small}
\end{center}
\end{table}

\paragraph{Reasoning:} One of the key strengths of the \methodname method is its ability to produce interpretable reasoning for the methods. To evaluate the quality of the reasoning, for 200 randomly sampled AudioCaps-Eval captions, we asked crowd-source workers (3 per caption) to rate three aspects of the generated scores on a 5-point Likert Scale: (1) How well the justification supported the score (Quality), (2) how fair the score was (Fairness), and (3) how well the score matched with the justification (Match). To provide a baseline, we employed \methodname with one of 36 variations of the justification ``No particular reason''. The results are given in \autoref{tab:reasoning}, where we found that the justifications both matched the score and were of significantly higher quality than the baselines $(p < 0.001)$. Further, we found that the justifications led humans to rate the score as more fair, with a significant $(p = 0.02)$ improvement over no justification (but the same score). 

\begin{table}
\caption{Human ratings of score/justification quality for \methodname (GPT-4o) on a subset of AudioCaps-Eval ($N=200$).}
\label{tab:reasoning}
\begin{center}
\setlength{\tabcolsep}{2.5pt}
\begin{tabularx}{\linewidth}{Xccccc}
\toprule
Measure & Fairness $(\uparrow)$ & Match $(\uparrow)$ & Quality $(\uparrow)$ \\
\midrule
FENSE & $2.97\pm1.02$ & - & - \\
\methodname/No Reason & $3.40\pm1.17$ & $2.92\pm1.29$ & $2.91\pm1.38$ \\
\methodname & $3.66\pm1.11$ & $3.80\pm1.03$ & $3.81\pm0.96$  \\ 
\bottomrule
\end{tabularx}
\end{center}
\end{table}

\paragraph{Qualitative Evaluations:} Some examples of the \methodname measure are given in \autoref{fig:qualitative}. In the first example, \methodname captures aggregate information in the set of baseline references and assigns a higher score to a caption that captures the entirety of that information, as opposed to closely matching a single caption. In the second, CLAIR-A penalizes for poor grammar, whereas other measures are fooled by high N-gram overlap. 

\paragraph{Discussion on the Cost of \methodname:} 
As discussed in \autoref{sec:method}, \methodname is designed to be computationally efficient compared to CLAIR. Unlike X-ACE \cite{wang2024x}, which requires multiple LLM calls per evaluation, \methodname processes each comparison with a single call. In practice, proprietary models complete evaluations in approximately 1.5 seconds per request at a cost of less than \$0.10, while open-source models such as Phi-3.5 achieve comparable performance in under 3 seconds on a single NVIDIA RTX 3090 GPU. As lightweight models like Phi-3 continue to improve, we expect \methodname to become even more cost-effective and widely applicable.
\section{Conclusion}

This paper introduces \methodname, a simple and interpretable domain-specific LLM-based measure for audio captioning. We demonstrate that not only is our simple approach well-aligned with human judgments, but also that such a method is significantly more interpretable to downstream human users. While \methodname is a first step towards LLM evaluation of audio captions, we hope that our work inspires further research into how LLMs can align with human judgment and can be used to develop simple and interpretable systems across a wide range of audio domains.

\section*{Acknowledgements}

As part of their affiliation with UC Berkeley, the authors were supported in part by the National Science Foundation, the Ford Foundation, and/or the Berkeley Artificial Intelligence Research (BAIR) Industrial Alliance program. Sky Computing Lab is supported by gifts from Accenture, AMD, Anyscale, Cisco, Google, IBM, Intel, Intesa Sanpaolo, Lambda, Lightspeed, Mibura, Microsoft, NVIDIA, Samsung SDS, and SAP. We thank Jeeweon Jung for his helpful contribution to the data used in the paper. GPT-4o was used to check the language of the paper (all sections) for spelling and grammar concerns.

\bibliographystyle{IEEEtranN}
\bibliography{mybib}

\end{document}